# Harnessing Artificial Intelligence for Wildlife Conservation

P. Fergus[1], C. Chalmers[1], S. Longmore[2], S. Wich[3]

[1]School of Computer Science and Mathematics, Liverpool John Moores Univeristy, Liverpool, L3 3AF.
[2]Astrophysics Research Institute, Liverpool John Moores University, L3 5RF.
[3]School of Biological and Environmental Sciences, Liverpool John Moores University, Liverpol, L3 3AF.

*Abstract*— **The rapid decline in global biodiversity demands innovative conservation strategies. This paper examines the use of artificial intelligence (AI) in wildlife conservation, focusing on the Conservation AI platform. Leveraging machine learning and computer vision, Conservation AI detects and classifies animals, humans, and poaching-related objects using visual spectrum and thermal infrared cameras. The platform processes this data with convolutional neural networks (CNNs) and Transformer architectures to monitor species, including those which are critically endangered. Real-time detection provides the immediate responses required for time-critical situations (e.g. poaching), while non-real-time analysis supports long-term wildlife monitoring and habitat health assessment. Case studies from Europe, North America, Africa, and Southeast Asia highlight the platform's success in species identification, biodiversity monitoring, and poaching prevention. The paper also discusses challenges related to data quality, model accuracy, and logistical constraints, while outlining future directions involving technological advancements, expansion into new geographical regions, and deeper collaboration with local communities and policymakers. Conservation AI represents a significant step forward in addressing the urgent challenges of wildlife conservation, offering a scalable and adaptable solution that can be implemented globally.**

*Index Terms*— **Artificial Intelligence (AI), Wildlife Conservation, Machine Learning, Species Identification, Poaching Prevention, Biodiversity Monitoring.**

## I. Introduction

The rapid decline in global biodiversity presents a profound threat to ecosystems and human well-being, underscoring the urgent need for more effective conservation strategies [1]. Traditional methods, while valuable, often fall short in addressing the scale and complexity of contemporary environmental challenges. Within this context, artificial intelligence (AI) has emerged as a transformative tool, offering new avenues for enhancing conservation efforts [2]. Conservation AI exemplifies this innovative application of AI in wildlife conservation. By leveraging advanced machine learning and computer vision technologies [3], Conservation AI seeks to detect and classify wildlife, monitor biodiversity, and support anti-poaching and other illegal activities [4], [5].

The platform, in collaboration with conservations partners around the world, employs a combination of visual spectrum and thermal infrared cameras, strategically deployed on camera traps and drones, to collect extensive data across various ecosystems. This data is processed using state-of-the-art AI models, including convolutional neural networks (CNNs) [6] and Transformer architectures [7], enabling the precise identification and tracking of animal species, particularly those at risk of extinction. The dual capability of real-time and non-real-time detection can potentially enhance the efficiency of conservation efforts, providing immediate response options as well as long-term monitoring solutions.

The integration of AI into conservation practices offers several distinct advantages [8]. Firstly, it facilitates continuous, non-invasive wildlife monitoring, thereby minimising human disturbance in sensitive habitats. Secondly, AI-driven analytics can swiftly process large datasets, uncovering patterns and trends that may elude human observation. Thirdly, AI's role in detecting poaching activities allows for rapid response and intervention, potentially averting the illegal hunting of endangered species [9].

Conservation AI's innovative approach is closely aligned with the needs of conservation organisations, research institutions, and local communities, ensuring that the technology is both effective and contextually appropriate. The reason this is so well aligned is that it has been developed in close collaboration with conservation groups. For instance, in South America, Conservation AI collaborates with local researchers to monitor jaguar populations and understand their habitat requirements. Similarly, in Africa, the platform has been developed alongside partners to track the movements of critically endangered species such as pangolins and bongos in Uganda and Kenya, contributing valuable insights to their preservation efforts [10]. These collaborations ensure that Conservation AI's technology is not only advanced but also context specific.

Furthermore, Conservation AI is committed to continuous improvement and adaptation. The platform regularly updates its AI models with new data and species information, incorporating user feedback to enhance its performance. This iterative approach ensures that Conservation AI remains at the forefront of technological advancements and is capable of addressing emerging conservation challenges. Additionally, the organisation invests in training and capacity-building programmes, empowering local communities and conservationists to effectively utilise AI tools.

This paper explores the methodologies employed by Conservation AI, its application across diverse conservation projects, and the outcomes achieved thus far. Through the examination of case studies and performance metrics, we aim to demonstrate the transformative potential of AI in supporting wildlife conservation and addressing the critical challenges faced by conservationists today.

## II. METHODOLOGIES

The methodology employed by Conservation AI is meticulously designed to harness advanced AI technologies for effective wildlife conservation. This section delineates the systematic approach adopted by Conservation AI, encompassing the entire process from data collection to the deployment of AI models for both real-time and non-real-time detection and classification. The methodology is structured to ensure thorough data acquisition, precise analysis, and the generation of actionable insights that directly inform conservation efforts. By integrating state-of-the-art machine learning techniques with practical field applications, Conservation AI seeks to significantly enhance the efficiency and effectiveness of conservation initiatives.

### A. Data Collection

Conservation AI employs a comprehensive and methodologically rigorous data collection strategy, utilising both camera traps and drones. By partnering with leading conservation organisations globally, including Chester Zoo, the Endangered Wildlife Trust, and the Greater Mahale Ecosystem Research and Conservation team, Conservation AI is able to amass a diverse array of datasets. This carefully crafted dataset creation process is what distinguishes Conservation AI from other organisations, ensuring that the data collected is not only extensive but also highly relevant and precise, thereby enhancing the effectiveness of the AI models used in conservation efforts.

The collected data encompasses both visual spectrum and thermal infrared imagery, including still images and video recordings of wildlife in their natural habitats. Camera trap data provides critical insights into species presence and behaviour within specific locales, while drone-acquired data expands observational capabilities to cover larger and more remote regions. This dual-method approach facilitates the capture of comprehensive wildlife activity patterns, encompassing both diurnal and nocturnal behaviours.

These meticulously curated datasets are utilised by Conservation AI to develop region-specific models tailored to various regions, such as Sub-Saharan Africa, the Americas, Asia, the UK, and Europe. Specialists in data collection, filtering, and quality control manage the data to optimise model performance. The deployment of a robust data processing pipeline ensures the production of high-quality datasets, which are essential for training AI models that are both accurate and reliable (repeatable).

### B. AI Models

The foundation of Conservation AI's technology lies in its sophisticated AI models, designed to perform complex image recognition tasks with a high degree of accuracy. Central to these models are CNNs [6] and Transformer architectures [7], both of which have proven exceptionally effective in the domain of computer vision. CNNs are particularly well-suited for processing grid-like data, such as images, excelling at identifying patterns, edges, and textures. Conversely, Transformer models are adept at capturing long-range dependencies and contextual relationships within data, making them powerful tools for interpreting more complex visual information.

Conservation AI's models are developed using TensorFlow [11] and PyTorch [12], two of the most widely used open-source machine learning frameworks, developed by Google and Meta, respectively. These frameworks provide robust and flexible platforms for designing, training, and deploying machine learning models at scale. The training process is both intensive and iterative, involving the ingestion of vast datasets comprising thousands of labelled images. These images represent a wide variety of species, human figures, and objects characteristic of typical animal behaviours, as well as potential indicators of poaching activities.

During training, the AI models are exposed to these labelled datasets, allowing them to progressively learn the distinguishing features and patterns associated with each category. This process, known as supervised learning, enables the models to continually improve their accuracy in detecting and classifying entities within new, unseen data. The models not only learn to identify different species and human presence but also recognise specific behaviours and anomalies that might suggest threats to wildlife, such as poaching or habitat encroachment [13].

What sets Conservation AI apart from other approaches is its careful crafting and continual refinement of these AI models. Techniques such as data augmentation [14], transfer learning [15], and fine-tuning [16] are employed to maximise model effectiveness. Data augmentation systematically varies training images - through rotations, translations, and other transformations - to help the models generalise better to different scenarios. Transfer learning allows the AI to leverage pre-trained models, adapting them to the specific requirements of conservation tasks with reduced computational costs and time. Fine-tuning further refines these models, ensuring they are sensitive to the nuances of local ecosystems and species.

In addition to these technical advancements, Conservation AI continuously updates and enhances its models by incorporating new data and feedback from conservation partners. This iterative improvement process, which we term situated learning and precision modelling, ensures that the AI remains at the cutting edge of technology, providing reliable and actionable insights that support global conservation efforts.

### C. Detection and Classification

Once trained, the AI models are deployed to process the data collected by the camera traps and drones. The detection and classification process involves several key steps:

1. Preprocessing: Raw images and videos undergo preprocessing to enhance quality and reduce noise. This step includes resizing, rescaling, normalisation, and augmentation of the data.
2. Inference: The pre-processed data is then fed into the AI models for inference. The models analyse the images and videos to detect the presence of specific animal species, humans, and objects.
3. Classification: Detected entities are classified into predefined categories, such as species type, human presence, and potential indicators of poaching (for example cars and people). The classification results are subsequently stored in a database for further analysis.

Figures 1 through 4 demonstrate the robust detection and classification capabilities of Conservation AI's models across various challenging scenarios. Figure 1 illustrates one of our elephant detections using the Sub-Saharan Africa model. This detection was made by a real-time camera installed in the Welgevonden Game Reserve in Limpopo Province, South Africa.

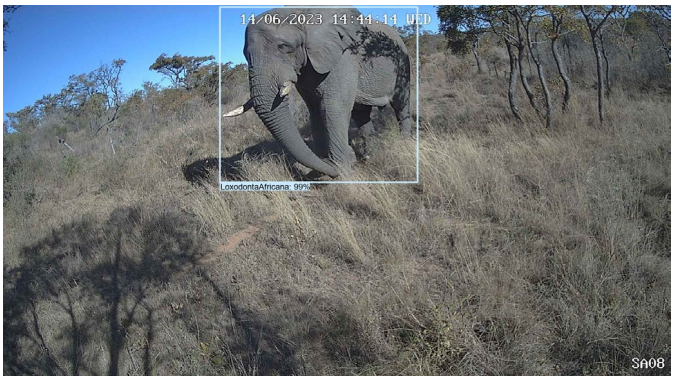

**Figure 1**. African elephant detected by a real-time camera in Limpopo in South Africa.

Figure 2 highlights a particularly challenging detection, which is central to the mission of Conservation AI. Captured by a real-time camera, it shows a zebra at night, detected from a considerable distance.

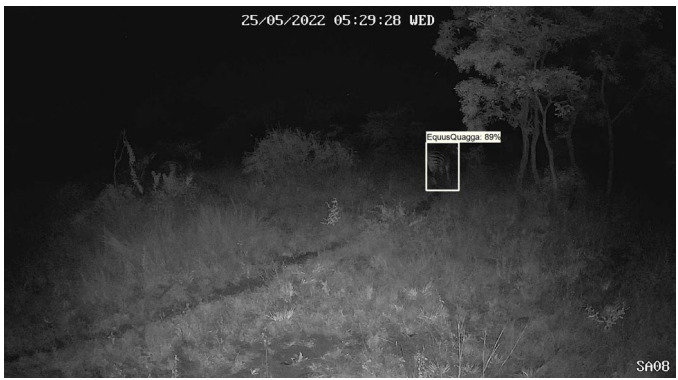

**Figure 2**. Zebra detected by a real-time camera in Welgevonden Game Reserve.

Figure 3 demonstrates the models' proficiency in handling occlusion and heavily camouflaged animals. In this example, a deer is partially concealed behind tree branches in a forest, yet the model successfully identifies the animal. Such capabilities are crucial for comprehensive biodiversity assessments and conservation efforts. This image was processed from data uploaded by a non-real-time camera.

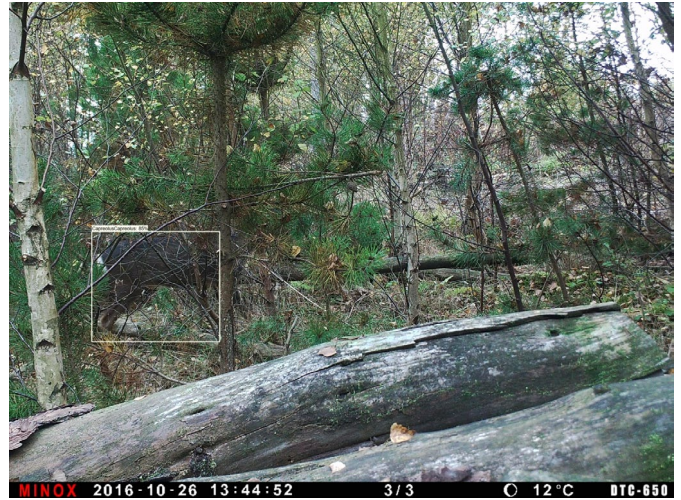

**Figure 3**. A deer situated in a wooded forest captured using a traditional non-real-time camera trap.

Figure 4 presents an intricate case, where our UK Mammals model not only detects a squirrel but accurately distinguishes it as a grey squirrel. This detection is particularly noteworthy given the challenging conditions: the animal is small, distant, and partially obscured by a tree, with poor lighting in the early morning.

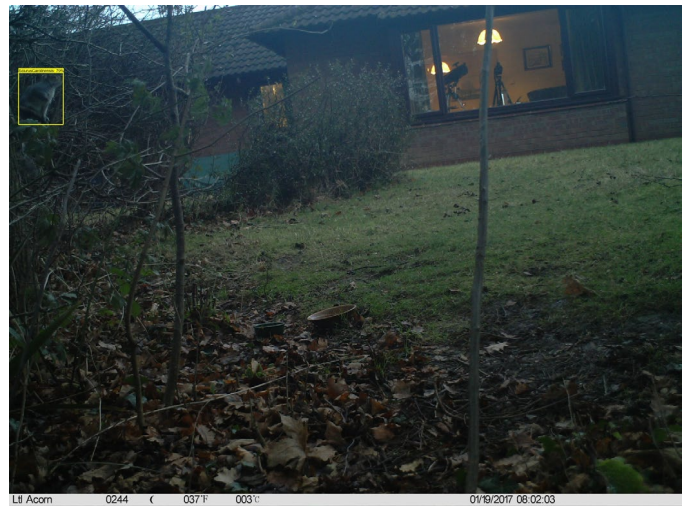

**Figure 4**. A grey squirrel detection in the early morning.

Images like these form the foundation of the datasets used by Conservation AI to train our models. When models are designed for camera traps, we exclusively use data from camera traps. The datasets reflect region-specific variables such as seasonality, day and night cycles, and varying weather conditions like rain and sunshine. Through the process of situated learning and precision modelling, our models undergo continuous fine-tuning, often taking up to a year to achieve the desired accuracy using data collected from the cameras we deploy. This ongoing learning and adaptation process is a unique feature of Conservation AI. Our commitment to developing long-term relationships with users and partners ensures that the models achieve the level of accuracy necessary for rigorous conservation studies.

### D. Real-Time and Non-Real-Time Capabilities

Conservation AI offers both real-time and non-real-time detection capabilities, each serving distinct but complementary roles in conservation efforts. Real-time detection is critical for enabling immediate responses to poaching activities [17]. The AI models process data on-the-fly, triggering alerts to conservationists and authorities when suspicious activities are identified. Figure 5 illustrates a real-time Conservation AI camera deployment in the UK.

Figure 5. Real-time camera trap deployed in the UK.

Non-real-time detection, on the other hand, involves batch processing of data collected over a specific period [18]. This approach typically utilises traditional camera traps that store data on SD cards, which must be retrieved at designated intervals for offline processing. This method is particularly valuable for long-term monitoring and the comprehensive analysis of wildlife populations and habitat health. By offering these dual capabilities, Conservation AI ensures that both urgent, time-sensitive conservation needs, and broader, long-term ecological assessments are effectively addressed.

### E. Data Management and Analysis

The classified data is securely stored in an online database, accessible to conservationists and researchers for further analysis. Conservation AI offers a suite of tools in a desktop client application, enabling conservationists to upload data, classify it, download the results, and generate reports for statistical analyses (see Figure 6).

Figure 6. Conservation AI desktop application for data processing and offline analytics.

These tools are instrumental in understanding trends, identifying poaching hotspots, and making informed decisions to shape conservation strategies. To ensure high performance and reliability, we utilise 3PAR flash storage for our data processing needs. Additionally, all data is backed up using a Synology NAS system, providing an extra layer of security and redundancy to safeguard this valuable information [19]. To further enhance our analytical capabilities, new reporting features are being developed based on Large Language Models (LLMs), which will improve our ability to generate insightful and comprehensive reports [7]. This robust data management and analysis infrastructure distinguishes Conservation AI from other approaches, ensuring that our partners have access to reliable, secure, and actionable information.

## III. APPLICATIONS AND CASE STUDIES

The practical applications of Conservation AI encompass a broad spectrum of conservation activities, including species identification, biodiversity monitoring, poaching prevention, and habitat restoration. As of this writing, Conservation AI supports over 900 active projects globally and processes more than 1.5 million images per week (and this is growing). By leveraging AI-driven insights, Conservation AI has significantly contributed to the enhancement and success of these diverse conservation initiatives worldwide. This section presents a selection of notable case studies that demonstrate the platform's effectiveness in real-world scenarios. These examples underscore the wide-ranging ways in which Conservation AI is utilised to address critical conservation challenges and protect endangered species, further distinguishing it as a leader in the field of wildlife conservation.

### A. Species Identification

One of the primary applications of Conservation AI is the automatic identification of wildlife species. By analysing images and videos captured by camera traps and drones, the AI models can accurately identify a wide range of species, including many that are threatened. This capability is particularly valuable in biodiversity hotspots where multiple species coexist. For example, in projects conducted across Sub-Saharan Africa, Conservation AI has successfully identified over 30 different species, providing essential data for conservationists to monitor and protect these animals [4]. This model has been effectively deployed in Uganda to monitor pangolins, in Kenya to track bongos, in the Maasai Mara to observe elephants, and in South Africa to safeguard black rhinos. Readers have already seen examples of this species identification capability in action, as illustrated in Figures 1 to 4, where the AI successfully identified an African elephant, a zebra, a deer, and a grey squirrel under challenging conditions. Figure 7 shows the detection of elephants in images captured from a consumer drone.

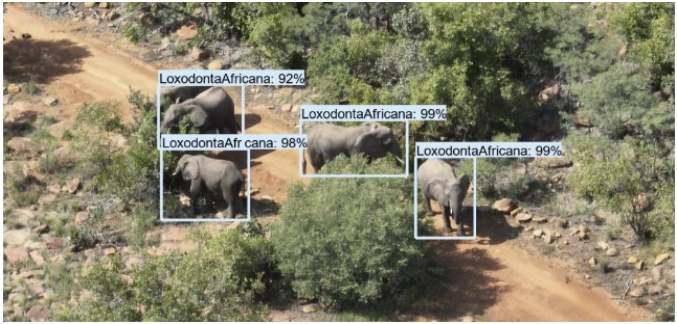

**Figure 7**. African elephant detected from a DJI Mavic 3 drone in Welgevonden Game Reserve in South Africa.

These examples underscore the platform's precision and reliability in species identification, making it a powerful tool for conservation efforts.

*B. Biodiversity Monitoring*

Conservation AI plays a pivotal role in monitoring biodiversity and assessing habitat health. By continuously collecting and analysing data, the platform enables conservationists to gain a deeper understanding of wildlife population dynamics and their interactions with the environment. For example, in a case study from Mexico, the South American Mammals Model is being utilised to monitor and track jaguars, while in California, the North American Mammals model is used to observe the movements and behaviours of mountain lions. The data collected through these initiatives has provided critical insights into migration patterns, feeding habits, and the impact of human-wildlife conflict. These findings not only enhance our understanding of these species but also inform strategies for mitigating threats and promoting coexistence between human and wildlife populations.

*C. Poaching Prevention*

One of the most impactful applications of Conservation AI is to support anti-poaching activities. The platform's real-time detection capabilities facilitate rapid responses to illegal hunting activities. In notable case studies from Uganda and the UK, Conservation AI successfully detected poaching activities involving pangolins and badgers, leading to convictions and prison sentences. The AI models identified suspicious activities within restricted areas, promptly sending alerts to park rangers, who were able to intervene and notify law enforcement. These interventions demonstrate the effectiveness of Conservation AI in supporting anti-poaching activities, underscoring its critical role in protecting endangered species.

*D. Community Engagement*

Engaging local communities is vital for the success of conservation projects. Conservation AI supports this engagement by providing accessible data and visualisations that can be shared with community members. In a project in India, the platform was employed to involve local communities in monitoring tiger populations. The data collected by Conservation AI was shared with villagers, who were trained to use the platform and actively contribute to conservation efforts. This collaborative approach not only enhanced data collection but also fostered a sense of ownership and responsibility among the community, empowering them to take an active role in protecting their local wildlife.

## IV. RESULTS AND DISCUSSION

The results from various projects using Conservation AI underscore the platform's effectiveness in advancing wildlife conservation efforts. Key performance metrics, such as the number of files processed, observations recorded, and species detected, highlight its impact. As of this writing, the platform has processed over 30 million images and identified more than 9 million animals across 88 species (please visit **www.conservationai.co.uk** to see updated stats). The platform's accuracy in species identification remains consistently high, achieving an average precision rate of 95%. These metrics illustrate the robustness and reliability of Conservation AI in supporting conservation initiatives worldwide.

*A. Model Performance*

The performance of the training results is evaluated using the mean average precision (mAP) with an intersection over union (IoU) threshold of 0.5 [20]. mAP is a widely used metric for assessing the effectiveness of object detection models, ranging from 0 to 1, with higher values indicating superior performance. Specifically, mAP@0.5 refers to the model's predictions being evaluated at an IoU threshold of 0.5, a standard measure in the field.

The Sub-Saharan Africa Mammals model is one of our most utilised models. Figure 8 displays the precision-recall curve for the individual classes (29 classes) within this model. The dark blue, thicker line represents the combined class, with a mAP@0.5 of 0.974. For clarity, we have omitted the class legend due to the large number of classes, which would make the colours and text difficult to discern.

An mAP@0.5 of 0.974 is considered excellent in the field of object detection. Achieving this level of precision indicates that

the model is highly accurate in detecting and classifying objects, making it exceptionally reliable for use in conservation studies.

while the Central Asian Mammals model (6 classes) achieves an even higher mAP@0.5 of 0.989 (Figure 12).

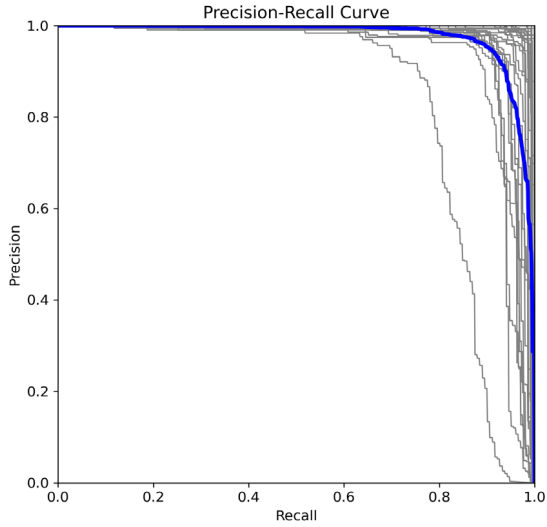

Figure 8. Precision-Recall Curve for the Sub-Saharan Africa Mammals model with an average 0.974 mAP@0.5.

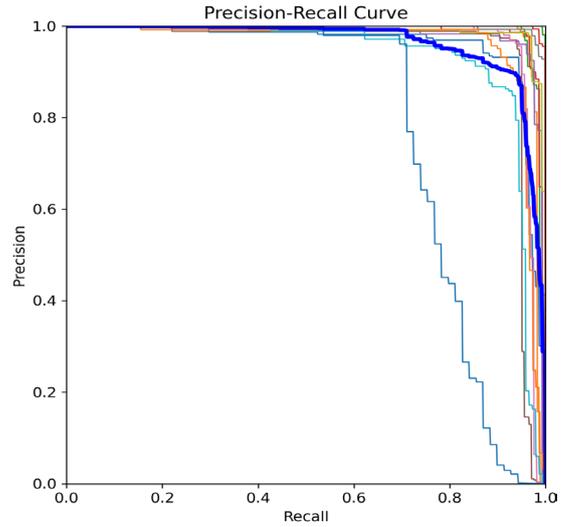

Figure 10. Precision-Recall Curve for the North American Mammals model with an average 0.961 mAP@0.5.

Figure 9 presents the same metric for the UK Mammals model, which includes 26 classes. The thick blue line indicates that a mAP@0.5 of 0.976 was achieved. By employing the same rigorous data preprocessing, data variance, and model training pipeline as used in the Sub-Saharan Africa Mammals model, we obtained similarly high results.

### B. Results from Use Cases

Several papers have been published demonstrating the capabilities of Conservation AI across various projects, with many more expected to follow. Appendix A provides a collection of images generated by our models in some of those studies, offering readers a glimpse into the wide-ranging applications of Conservation AI. These images illustrate the challenging environments in which our models operate and highlight some of the complex detections that can be difficult even for humans to classify. For more detailed discussions on specific models and studies, readers are directed to the following collection of papers [4], [5], [10], [21], [22], [23], [24], [25], [26], [27], [28].

### C. Challenges and Limitations

Despite its successes, Conservation AI faces several challenges and limitations. One of the primary challenges lies in the quality of data collected. Camera traps, more so than drones (although they have their own problems, i.e. distance and scale), often capture low-quality images and videos under a wide range of adverse weather and environmental conditions. Acquiring sufficient region-specific data to effectively train our models - especially when dealing with rare or critically endangered animals - can be particularly difficult. This limitation can affect the accuracy of the AI models and, in some cases, slow down progress.

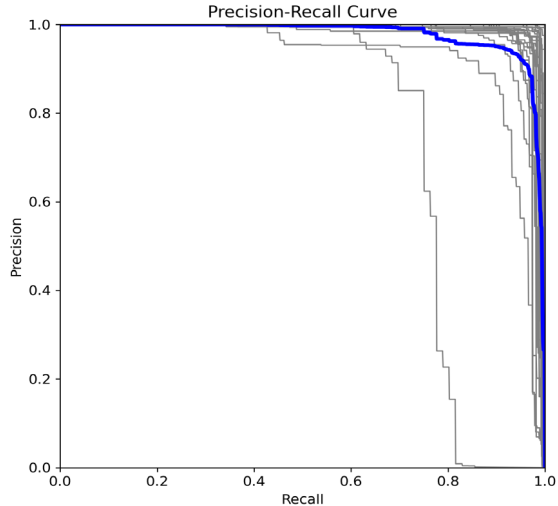

Figure 9. Precision-Recall Curve for the UK Mammals model with an average 0.976 mAP@0.5.

Consistent with the results from our Sub-Saharan Africa Mammals model, a mAP@0.5 of 0.976 is regarded as an excellent outcome. The North American Mammals model, which includes 12 classes, similarly demonstrates strong performance with a mAP@0.5 of 0.961, as shown in Figure 10. The Indomalayan Mammals model (10 classes) produces comparable results with a mAP@0.5 of 0.977 (Figure 11),

Even when data is available, the tagging process is time-consuming and crucial to the overall pipeline. If tagging is not performed meticulously, it can significantly degrade the quality of the models. Maintaining high standards in data management is both time-consuming and costly, and as a non-profit platform, ensuring the necessary data management capacity remains an ongoing challenge.

Additionally, deploying and maintaining camera traps and drones in remote areas presents logistical challenges and requires considerable resources. While transfer learning helps mitigate some data collection issues and contributes to our strong results, it does not eliminate the need to create high-quality, balanced datasets for model training. Features present in the deployed environment may not be included in the initial base models, necessitating the need to deploy hardware in model-specific regions to create custom datasets. This can be likened to overfitting models to regions of intended use.

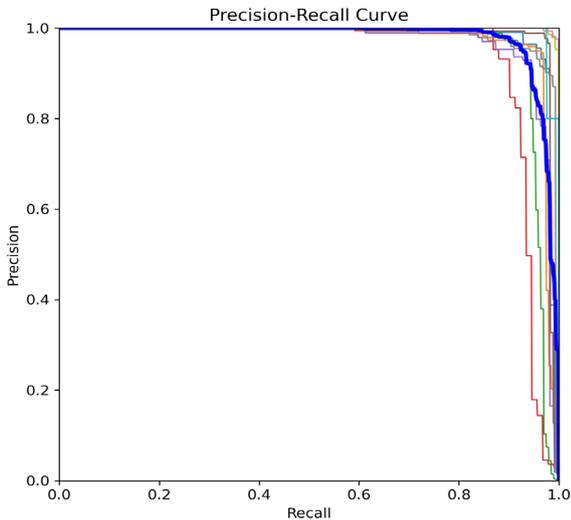

Figure 11. Precision-Recall Curve for the Indomalayan Mammals model with an average 0.977 mAP@0.5.

Another limitation is the need for continuous training and updating of AI models. As new species are added and environmental conditions evolve, base models must be retrained from scratch to maintain their accuracy. It is not sufficient to simply add new data or classes to an existing model, as transfer learning requires fine-tuning earlier layers in the network. Consequently, the entire dataset, including the new information, must be passed through the training cycle again, which demands significant computational resources - an expensive requirement over time.

Moreover, the deployment of AI technology in conservation efforts necessitates collaboration with local communities and authorities. This collaboration can be challenging due to cultural and logistical barriers, as well as general mistrust in AI solutions.

### D. Discussion

The results from the models presented, along with the images collected from various projects (as shown in Appendix A), demonstrate the significant potential of Conservation AI to revolutionise wildlife conservation. The platform's capacity to process vast amounts of data rapidly - benchmarking at 100 million images processed in seven days using 8 RTX A6000 GPUs loaded with 130 Faster RCNN deep learning vision models [29] - and with high accuracy, provides conservationists with critical support to conserve wildlife populations and their habitats. The real-time detection capabilities are particularly advantageous for preventing poaching and protecting endangered species, highlighting Conservation AI's role in advancing the effectiveness of conservation strategies.

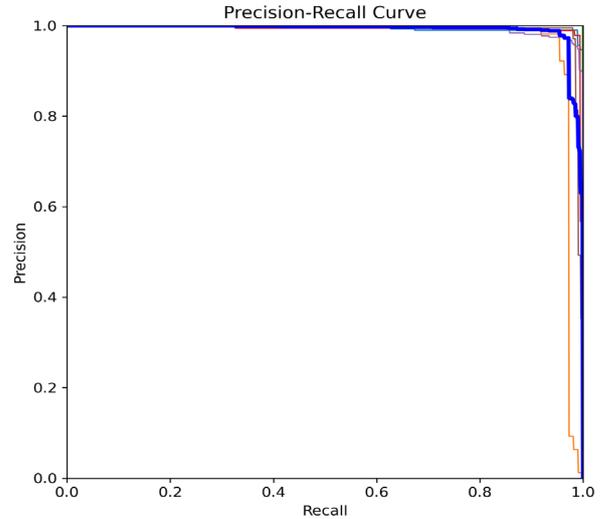

Figure 12. Precision-Recall Curve for the Central Asian Mammals model with an average 0.989 mAP@0.5.

However, the challenges and limitations emphasise the need for ongoing improvement and collaboration. Enhancing the quality of data collection, increasing the robustness of AI models, and fostering stronger partnerships with local communities and authorities are crucial for maximising the impact of Conservation AI.

## V. FUTURE DIRECTIONS

The future of Conservation AI is characterised by continuous innovation and expansion. This section outlines the technological advancements, expansion plans, and collaborative initiatives that will drive the platform's development. By focusing on refining AI models, extending geographical coverage, and strengthening partnerships with local communities and policymakers, Conservation AI aims to amplify its impact on wildlife conservation. Furthermore, ongoing research and development efforts will explore new applications of AI while addressing ethical considerations, ensuring that the technology is employed responsibly and effectively for the greater good of conservation.

### A. Technological Advances

A key area of focus is extending and enhancing the accuracy and robustness of AI models. This will be accomplished by incorporating new data and emerging deep learning models, particularly those with multimodal capabilities, such as vision transformers [30]. Vision transformers will enable more sophisticated and nuanced analysis by integrating diverse data types, improving the platform's ability to detect and classify wildlife and poaching activities. Although audio modelling is already part of the Conservation AI model ecosystem, fully integrating it with the visual components of the platform remains an ongoing goal. This integration will further strengthen the platform's capabilities in providing comprehensive conservation solutions [24].

Another promising direction for Conservation AI is the development of edge AI solutions. By deploying AI models directly on camera traps and drones, Conservation AI can process data locally, reducing the reliance on constant internet connectivity and enabling real-time decision-making even in remote areas. This approach also conserves bandwidth and energy, making the system more sustainable and efficient.

B. *Expansion Plans*

Conservation AI is committed to expanding its platform to support a broader range of devices and regions. This expansion includes integrating with various types of camera traps, drones, and other data collection tools. Current efforts are focused on abstracting the sensor hardware to create a single interoperable solution. A key feature common to camera traps and other types of edge device is the presence of an SD card slot for data storage. We intend to leverage this feature by using wireless WiFi SD cards capable of transmitting data to a base station, which can then relay it to our servers via wide area communications (4/5G where available, or STARLINK Mini where connectivity is limited) [17]. By extending geographical coverage, Conservation AI aims to support conservation efforts across diverse ecosystems, from tropical rainforests to arid deserts.

The platform also plans to collaborate more extensively with conservation organisations, research institutions, and governments. As our user base continues to expand, we are committed to promoting widespread adoption in tandem with the platform's growth. These partnerships will be crucial in scaling up the deployment of Conservation AI and ensuring that the technology is customised to meet the specific needs of diverse conservation projects. Furthermore, expanding the user base will contribute to the collection of more data, which in turn will enhance the accuracy and effectiveness of the AI models through ongoing training activities.

C. *Community Engagement*

Collaboration is essential to the success of Conservation AI. The platform is committed to fostering partnerships with local communities, conservationists, and policymakers. Engaging local communities in data collection and monitoring not only enhances the quality of data but also ensures that conservation efforts are culturally sensitive and sustainable. Training programmes and workshops can empower community members to utilise the technology effectively and contribute actively to conservation efforts. The Conservation AI team frequently travels to conservation sites to gain a deeper understanding of the specific challenges, current conservation methods, and determine how our platform can support study pipelines. This hands-on approach has enabled us to develop long-term relationships, which we intend to strengthen through increased outreach efforts.

Furthermore, collaboration with policymakers is crucial for creating supportive regulatory frameworks that facilitate the deployment of AI in conservation. This is especially important as bio credits [31] emerge as a key tool in biodiversity monitoring and management, alongside the adoption of more radical concepts such as interspecies money, where animals own their own resources [4]. By working together, stakeholders can address legal and ethical considerations, including data privacy and the impact of AI on local wildlife and communities [32].

D. *Research and Development*

Ongoing research and development are vital for the continuous enhancement of Conservation AI. Future research will explore new applications of AI in conservation, such as predicting the impact of climate change on wildlife populations and habitats. This involves extending beyond classification to focus on predictive modelling, utilising techniques similar to recurrent neural networks (RNNs) [33], long short-term memory (LSTM) networks [34], and the more recent advancements in Transformers [7].

Investing in research on the ethical implications of AI in conservation is equally important. This includes examining potential biases in AI models and ensuring that the technology is employed responsibly and transparently, avoiding any negative impacts on biodiversity. By addressing these critical issues, Conservation AI can build trust with stakeholders and ensure that its technology is used for the greater good.

VI. CONCLUSIONS

In this paper, we have explored the innovative application of artificial intelligence in wildlife conservation through the lens of Conservation AI. By leveraging advanced AI models and sophisticated data collection techniques, Conservation AI has demonstrated significant potential in enhancing conservation efforts. The platform's ability to accurately detect and classify wildlife, monitor biodiversity, and prevent poaching activities offers valuable tools for conservationists and researchers.

The case studies presented highlight the diverse applications of Conservation AI, ranging from species identification and biodiversity monitoring to poaching prevention and habitat restoration. These examples underscore the platform's effectiveness in addressing some of the most pressing challenges in wildlife conservation. However, the success of Conservation AI also reveals several challenges and limitations, including issues related to data quality, model accuracy, and logistical constraints.

Looking ahead, the future of Conservation AI lies in ongoing technological advancements, expansion plans, and collaborative initiatives. By further improving AI models, expanding geographical coverage, and fostering partnerships with local communities and policymakers, Conservation AI can amplify its impact on wildlife conservation. Continued research and development will be essential for exploring new applications of AI and addressing ethical considerations.

In conclusion, Conservation AI represents a significant step towards harnessing the power of artificial intelligence for the greater good of wildlife conservation. The integration of AI into conservation efforts not only enhances the efficiency and accuracy of monitoring and protection strategies but also opens new possibilities for understanding and preserving our planet's biodiversity. As we progress, it is crucial to continue investing in AI-driven conservation technologies and fostering collaboration among stakeholders to ensure a sustainable and thriving future for wildlife.

ACKNOWLEDGMENTS

In the early days of Conservation AI, we owe a significant debt of gratitude to Knowsley Safari, who provided invaluable support with data collection and model training by allowing us to install real-time cameras across their animal paddocks. We also thank them for their ongoing support, as our cameras continue to be deployed and collect data to this day. We would also like to extend our gratitude to Welgevonden Game Reserve in Limpopo Provence in South Africa who installed over 25 of our real-time cameras and evaluated the Sub-Saharan Africa model we developed with Knowsley Safari. We are also deeply indebted to Chester Zoo, a long-term collaborator and active partner in the development of Conservation AI. Their contributions, including extensive data provision and the facilitation of real-time camera installations in several African countries for ongoing studies - such as pangolin monitoring in Uganda and bongo monitoring in Kenya - have been instrumental in advancing our work. There are many more contributors who have helped develop Conservation AI – too many to mention in this paper – and we thank you all. Lastly, we would like to thank UKRI, particularly STFC and EPSRC, in the UK for providing valuable funds to develop Conservation AI. We would also like to thank the U.S. Fish and Wildlife Service for the funding and support they have provided us.

APPENDIX A

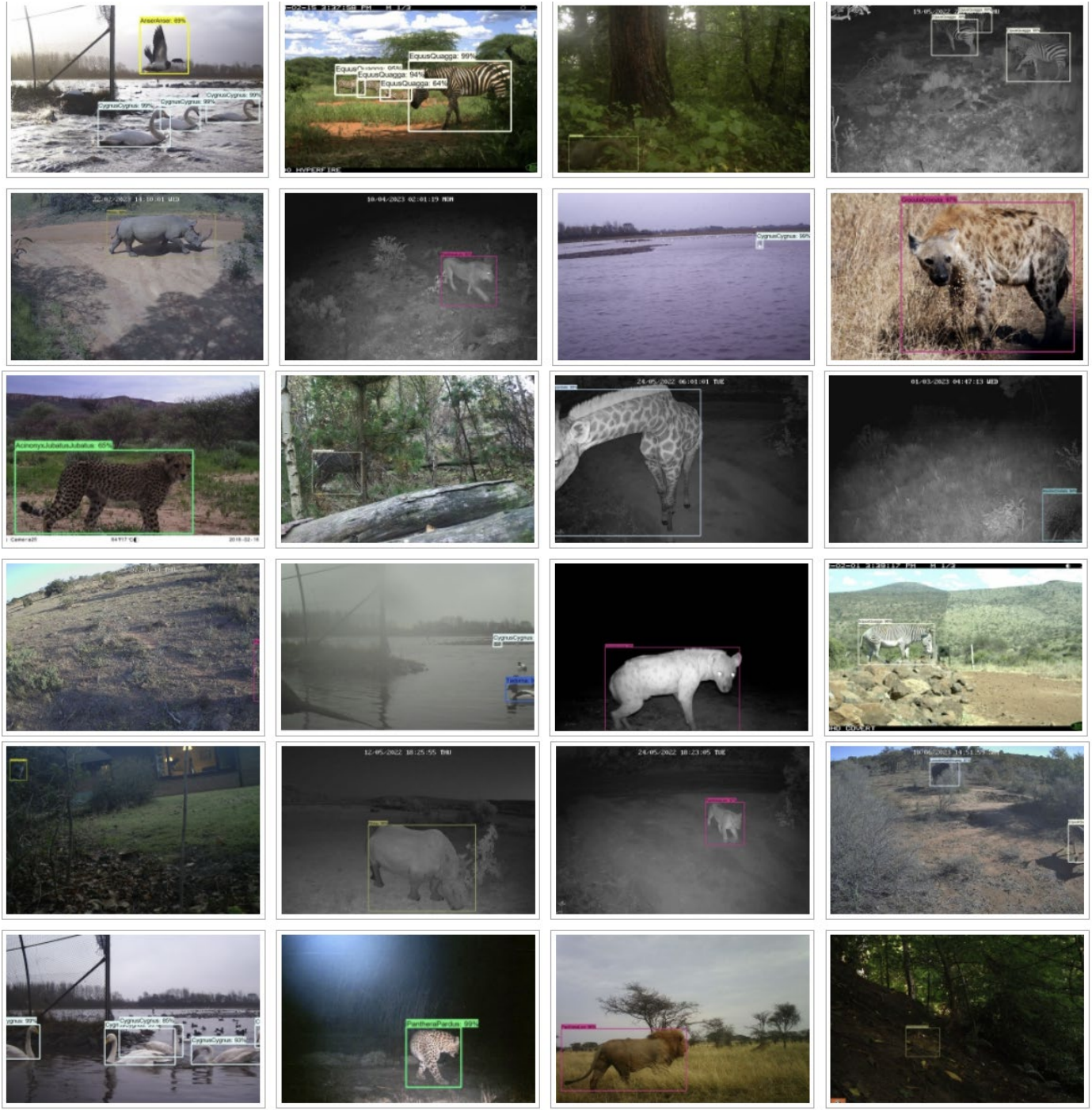

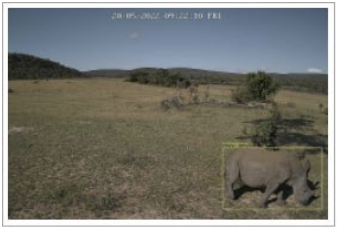 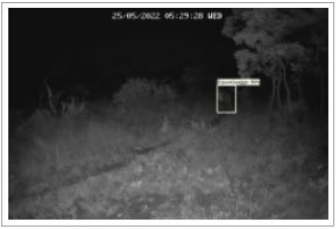 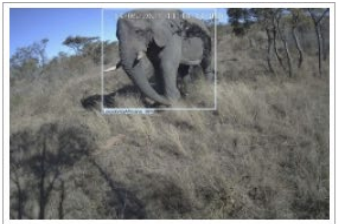 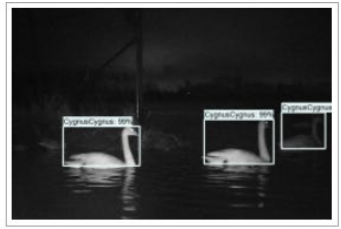
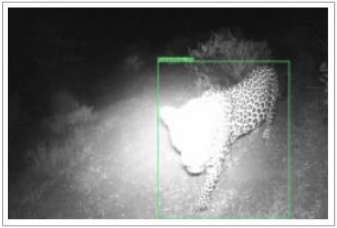 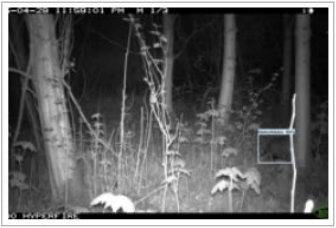 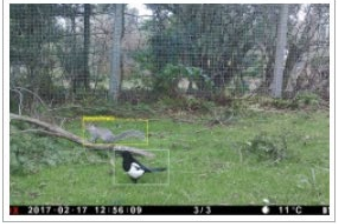 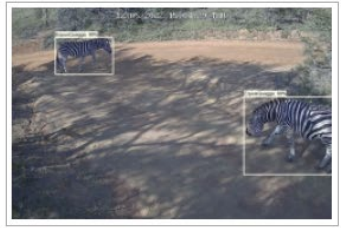
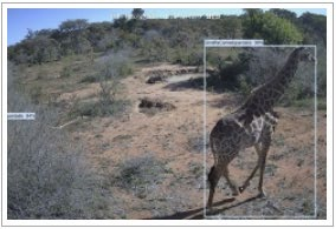 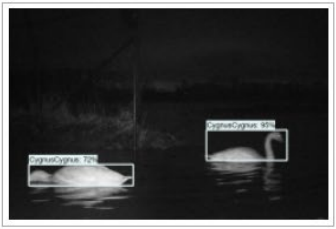 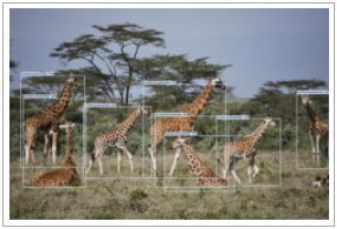 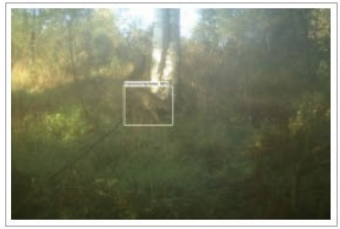
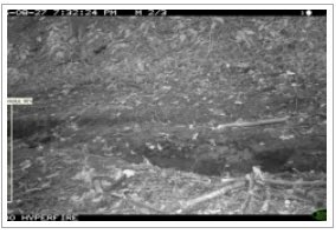 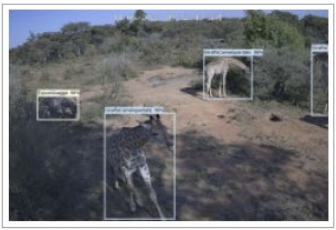 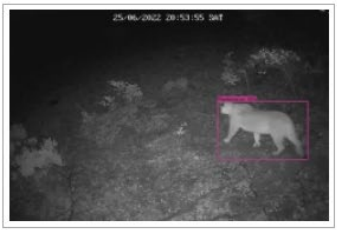 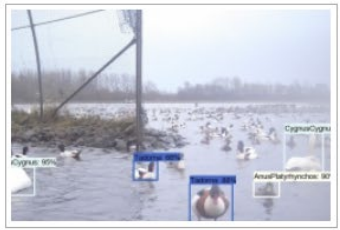
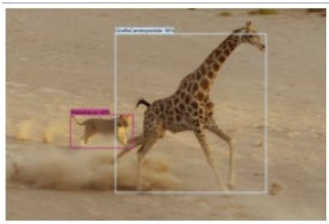 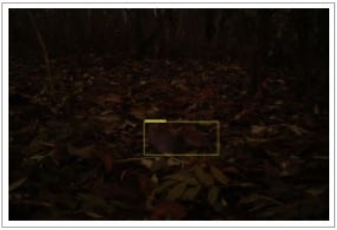 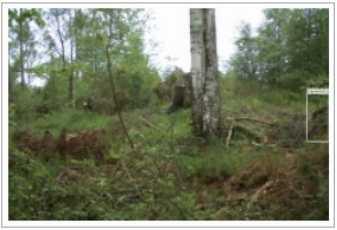 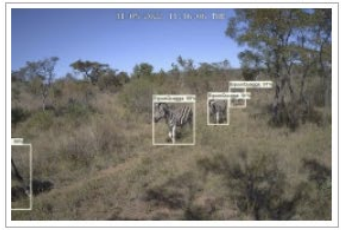
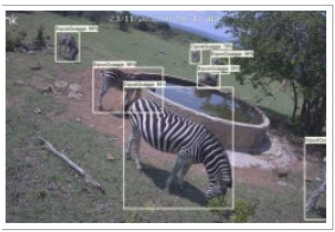 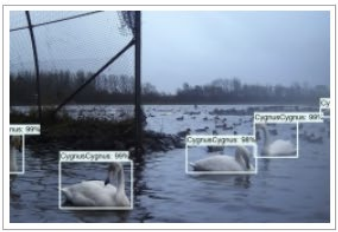 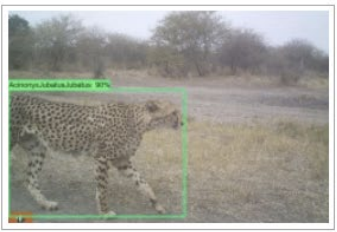 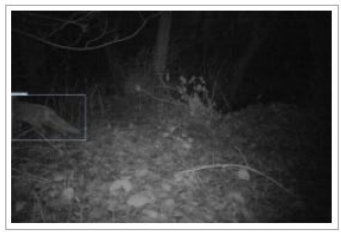
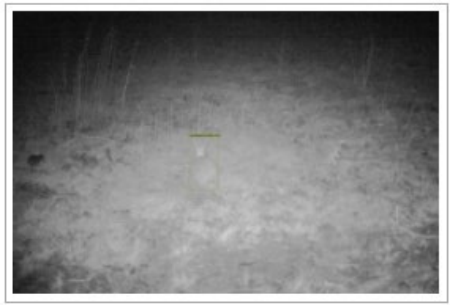 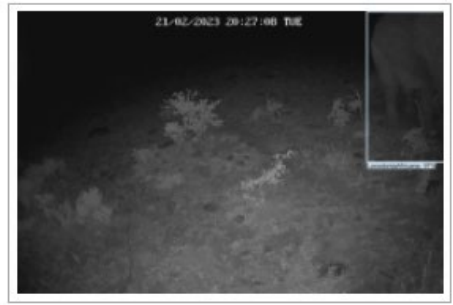 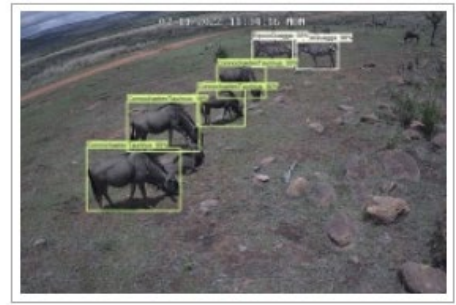

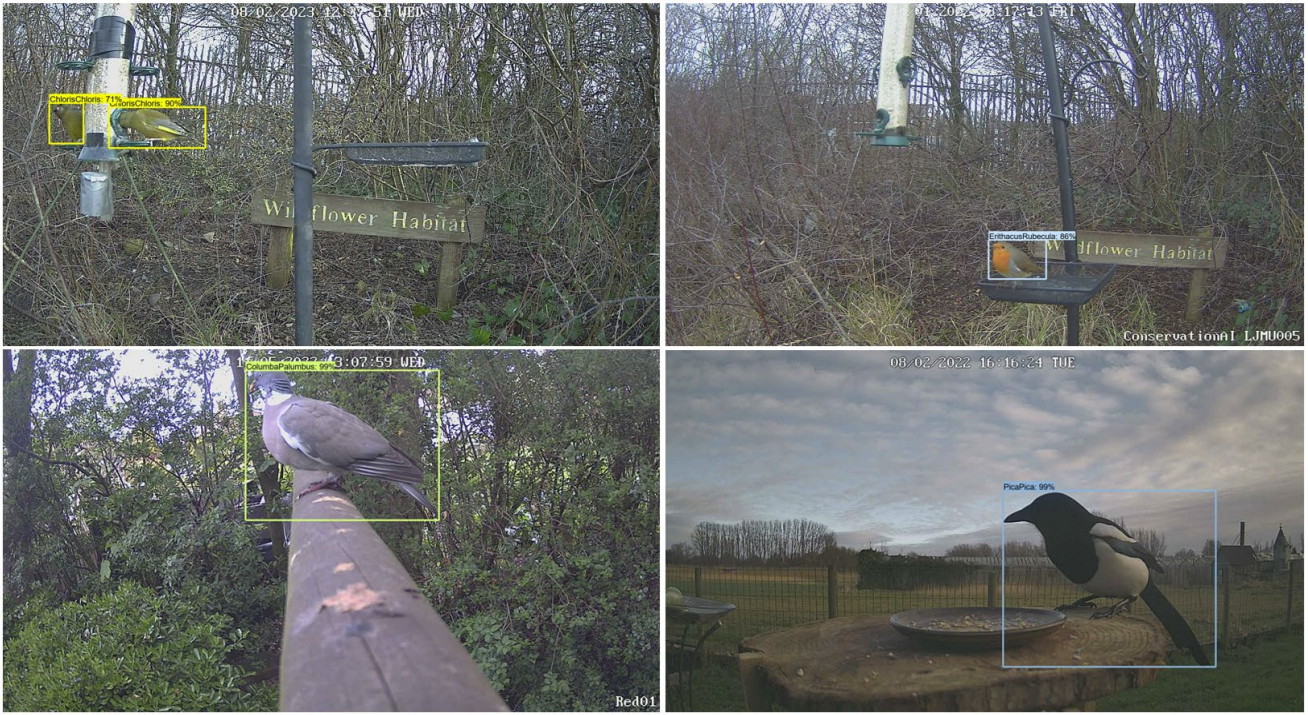

Figure 13: Collection of detections from several of our studies conducted in different geographical locations globally